\documentclass[letterpaper]{article} 
\usepackage{aaai25}  
\usepackage{times}  
\usepackage{helvet}  
\usepackage{courier}  
\usepackage[hyphens]{url}  
\usepackage{graphicx} 
\usepackage{natbib}  
\usepackage{caption} 
\usepackage{algorithm}
\usepackage{algorithmic}
\usepackage[table,xcdraw]{xcolor}
\usepackage{colortbl}
\usepackage{amsmath,amsfonts}
\usepackage{booktabs}
\usepackage{multirow}
\usepackage{color}
\usepackage[symbol]{footmisc}

\newcommand{\prettyblue}{\color[RGB]{59,87,234}}
\newcommand{\hanx}[1]{{\prettyblue[\textbf{Hanx:} #1]}}
\newcommand{\dgreen}{\color[rgb]{0.25,0.875,0.8125}}
\newcommand{\liu}[1]{{\dgreen[\textbf{Liu:} #1]}}
\newcommand{\xz}[1]{{\color[rgb]{1,0,0}#1}}

\usepackage{newfloat}
\usepackage{listings}
\DeclareCaptionStyle{ruled}{labelfont=normalfont,labelsep=colon,strut=off} 
\floatstyle{ruled}
\newfloat{listing}{tb}{lst}{}
\floatname{listing}{Listing}
\title{RIDE: Boosting 3D Object Detection for LiDAR Point Clouds via Rotation-Invariant Analysis}

\author {
    Zhaoxuan Wang\textsuperscript{\rm },
    Xu Han\textsuperscript{\rm },
    Hongxin Liu\textsuperscript{\rm },
    Xianzhi Li\textsuperscript{\rm }
}
\affiliations {
    \textsuperscript{\rm }Huazhong University of Science and Technology\\
    \{wang\_zx, xhanxu, hongxin\_liu, xzli\}@hust.edu.cn
}

\usepackage[switch]{lineno}
\begin{document}


\maketitle

\begin{abstract}
The rotation robustness property has drawn much attention to point cloud analysis, whereas it still poses a critical challenge in 3D object detection. When subjected to arbitrary rotation, most existing detectors fail to produce expected outputs due to the poor rotation robustness. 
In this paper, we present \textbf{RIDE}, a pioneering exploration of \textbf{R}otation-\textbf{I}nvariance for the 3D LiDAR-point-based object \textbf{DE}tector, 
with the key idea of designing rotation-invariant features from LiDAR scenes and then effectively incorporating them into existing 3D detectors.
Specifically, we design a bi-feature extractor that extracts (i) object-aware features though sensitive to rotation but preserve geometry well, and (ii) rotation-invariant features, which lose geometric information to a certain extent but are robust to rotation. 
These two kinds of features complement each other to decode 3D proposals that are robust to arbitrary rotations.
Particularly, our RIDE is compatible and easy to plug into the existing one-stage and two-stage 3D detectors, and boosts both detection performance and rotation robustness. Extensive experiments on the standard benchmarks showcase that the mean average precision (mAP) and rotation robustness can be significantly boosted by integrating with our RIDE, with \textbf{+5.6\%} mAP and \textbf{53\%} rotation robustness improvement on KITTI, \textbf{+5.1\%} and \textbf{28\%} improvement correspondingly on nuScenes. The code will be available soon.
\end{abstract}

\section{Introduction}
\label{sec:intro}

Autonomous driving, being able to navigate without human intervention, is no longer a distant vision of the future.
As an indispensable component of automotive perception system, 3D object detection aims to predict the locations, sizes, and classes of critical objects, e.g., cars, pedestrians, and cyclists, near an autonomous vehicle.
3D object detection methods have evolved rapidly with the advances in deep learning techniques.
In recent years, to address the inevitable quantization loss of voxel-based 3D object detectors \cite{yan2018second, lang2019pointpillars, deng2021voxel}, 3D object detection based on LiDAR point clouds \cite{shi2019pointrcnn, yang20203dssd, zhang2022not} are proposed.
However, none of them considers the stability of detection results when the inputs are rotated.

It is quite natural that the orientation of the same object towards LiDAR might vary, thus causing the scanned point coordinates to change, although the object itself is not changed.
Existing point-based detectors take raw 3D coordinates as network input for extracting latent features, thus causing the detectors sensitive to rotation.
To improve the rotation robustness for 3D object detection, EON \cite{yu2022rotationally} and TED \cite{wu2023transformation} 
are developed with the key idea of achieving rotation equivariance by assuming a list of pre-defined rotation angles, but would result in a performance drop when testing with arbitrary rotation angles.



\begin{figure}[t]
	\centering
	\includegraphics[width=\linewidth]{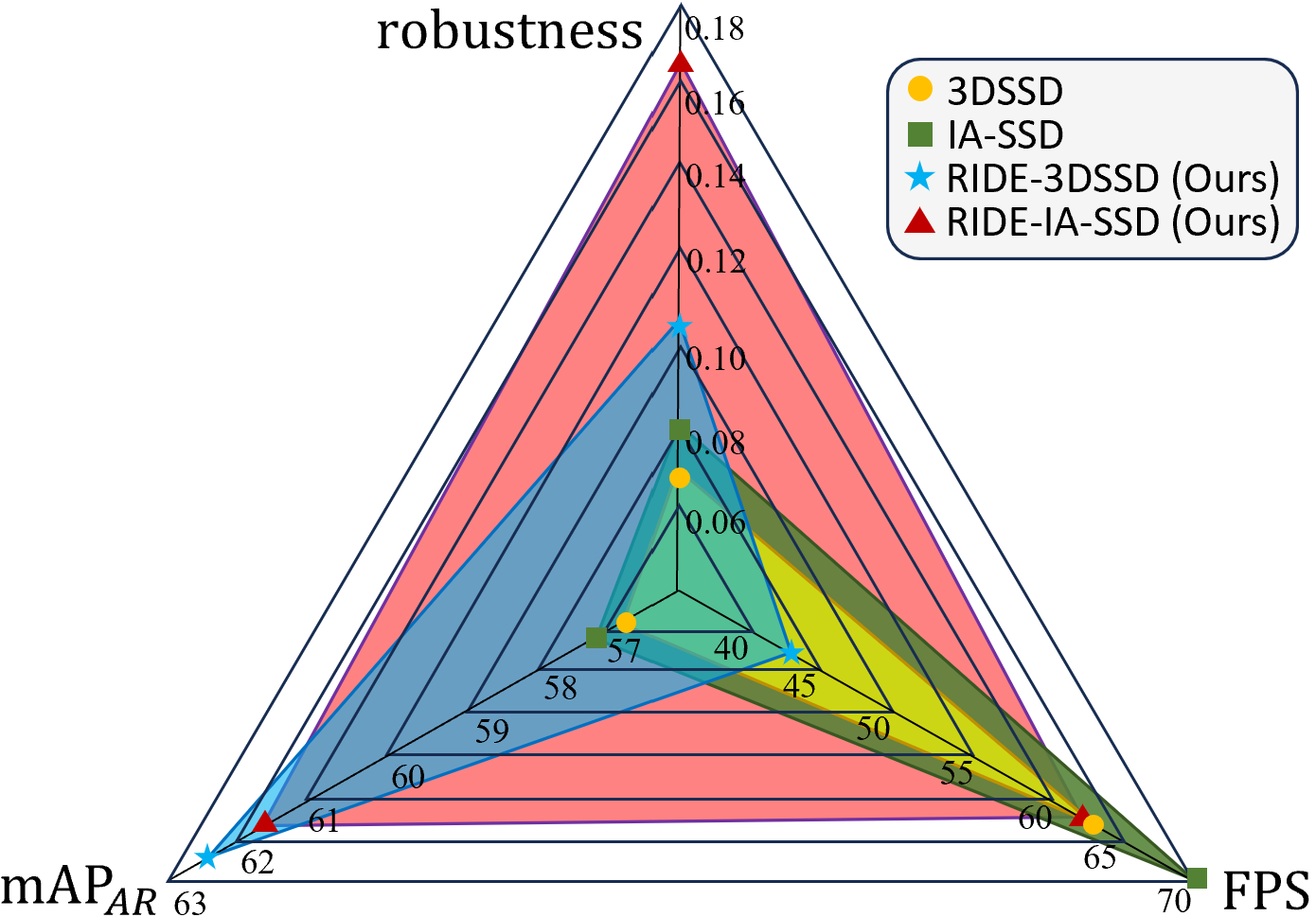}
	\caption{Compare the detection performance, rotation robustness and inference speed between the original 3D detectors (i.e., 3DSSD \cite{yang20203dssd} and IA-SSD \cite{zhang2022not}) and the associated ones equipped with our RIDE.
    Clearly, our RIDE further boosts both perception precision and rotation robustness with an acceptable decrease in speed; see the red and blue regions.
 }
	\label{fig:teaser}
\end{figure}

Inspired by the rotation-invariant analysis on the point cloud \cite{zhang2022riconv++, chen2022devil} that achieves consistent performance under arbitrary rotation, we pioneer the investigation of rotation-invariance on the LiDAR-point-based 3D detection, aiming to enhance the rotation robustness for the existing detectors.
Note that, it is not straightforward to transplant existing rotation-invariant techniques into 3D object detection due to the following reasons.
First, existing rotation-invariant methods are all developed for a single 3D shape, rather than a LiDAR-based sparse scene.
Second, to solve the rotation sensitivity, existing methods choose to completely replace the point coordinates with rotation-invariant features as network input. This simple and crude way will lead to a large amount of information loss, making the network unable to effectively detect objects from large scenes.

In our work, we propose \textbf{RIDE}, the \emph{first} attempt to explore \textbf{R}otation-\textbf{I}nvariance for 3D \textbf{DE}tector from LiDAR point clouds. Specifically, we design a simple but quite effective module termed rotation-invariance block (RIB) for extracting pure rotation-invariant features (RIFs) from input LiDAR point clouds that can be flexibly incorporated into various existing point-based detector encoders. On the other hand, to avoid much information loss, when equipping our RIB to existing detectors, we still keep the original rotation-sensitive feature extractor to obtain object-aware features (OAFs). Therefore, our method can extract both RIFs and OAFs at the same layer, termed the bi-set abstraction (Bi-SA) layer. Next, we stack multiple Bi-SA layers to form a novel bi-feature extractor, then followed by a detection head to regress 3D proposals. 

Intuitively, compared to existing 3D detectors, the features obtained from our RIDE not only contain rich geometric information, but also are robust to rotations.
Meanwhile, Compared to rotation-robust detectors EON \cite{yu2022rotationally} and TED \cite{wu2023transformation} that explore the rotation-equivariance property for the 3D object detection via a few discrete rotation angles, our method can tackle any rotation situation, which demonstrates better robustness to the arbitrary and unseen rotation angles. Particularly, our RIDE can be easily incorporated into existing state-of-the-art detectors without any restrictions.

To evaluate the detection performance as well as the rotation robustness, we conduct experiments on two popular benchmark datasets KITTI \cite{geiger2012we} and nuScenes \cite{caesar2020nuscenes} under two cases: the input is randomly rotated with (1) default angles (i.e., $\in \left[ -\pi /4,\pi /4 \right]$) and (2) arbitrary angles (i.e., $\in \left[-\pi, \pi\right]$), so that rotation robustness can be calculated by the difference in these two cases.
Extensive experiments on the KITTI showcase that detectors equipping our RIDE achieve \textbf{+1.3\%} and \textbf{+5.6\%} mean average precision (mAP) boost under the test with default rotation (\textit{DR}) and arbitrary rotation (\textit{AR}) cases respectively compared to the baseline, and obtain \textbf{53\%} improvement in rotation robustness. On the nuScenes, we bring \textbf{+4.2\%} and \textbf{+5.1\%} mAP boost under \textit{DR} and \textit{AR} cases respectively, and improve rotation robustness by \textbf{28\%}. 

Figure \ref{fig:teaser} demonstrates an example comparison result on the KITTI. Clearly, when equipping our RIDE to the existing two baselines (see the red and blue regions), both the perception precision and rotation robustness are significantly improved without an obvious decrease in speed.
The Experiments Section provides more detailed results.

\section{Related Works}
\label{sec:rw}

\paragraph{Point-based methods for 3D object detection.} 
To avoid the quantization loss caused by transforming point clouds to voxels \cite{yan2018second, lang2019pointpillars, shi2020pv, deng2021voxel}, a pioneer work PointRCNN \cite{shi2019pointrcnn} was proposed to directly process on 3D LiDAR point clouds for generating 3D proposals.
As a two-stage framework, PointRCNN first employs a PointNet++-based encoder to extract first-stage object proposals, which are then refined by exploiting semantic features and local information in the second stage.
%
To alleviate the time consumption caused by the second-stage refinement,
VoteNet \cite{qi2019deep} proposed a single-stage detection (SSD) framework using the deep Hough voting to predict the centroid point of objects. Inspired by this voting mechanism, 3DSSD \cite{yang20203dssd} designed an SSD network with the feature-distance down-sampling strategy, which preserved distinct features while achieving promising performance. Zhang et al. proposed IA-SSD \cite{zhang2022not} with the instance-aware down-sampling strategy, further improving efficiency. SASA \cite{chen2022sasa} introduced a plug-in farthest point sampling strategy based on the semantic distance to efficiently choose foreground points.

Despite the promising performance the above methods have achieved, none of them considers the impact of the input rotation on the stability of the prediction.
That is to say, when the input scene points rotate, the detected 3D oriented bounding boxes (OBBs) of objects using the above methods may be unreliable, due to the change of objects’ orientation.
To improve the rotation robustness for 3D object detection, EON \cite{yu2022rotationally} and TED \cite{wu2023transformation} proposed rotation-equivariant methods for autonomous driving scenarios. Specifically, EON can be integrated with existing detectors with the design of rotation-equivariance suspension for retaining orientation, while aggregating invariant features to achieve object-level equivariance. TED is an efficient voxel-based detector, that extracts transformation-equivariant voxel features via a sparse convolution backbone, and then aligns and aggregates these features for proposal prediction and refinement.
However, they achieved rotation-equivariance assuming a list of pre-defined rotation angles, while this idea cannot cover all the rotation cases, thus leading to unstable prediction under unseen rotation angles. Additionally, they lacked validation of the equivariant performance exceeding the default rotation angles $ \in \left[ -\pi /4,\pi /4 \right] $. 
Instead of incorporating rotation-equivariance to the 3D object detection, we pioneer the investigation of rotation-invariance to assist robust feature extraction, so that the features extracted by the encoder are less sensitive to \emph{arbitrary} rotation, thereby making the detected 3D bounding boxes more accurate and stable.


\paragraph{Rotation-invariant methods for point cloud analysis.} 
Theoretically, the features extracted from a 3D object should be unchanged regardless of the rotations. Nonetheless, most works utilize the 3D point coordinates as input. Consequently, as the object undergoes rotation, the point coordinates inevitably shift, so the consistency of the features cannot be guaranteed.
To overcome this limitation, some methods \cite{zhang2019rotation, li2021rotation, zhang2022riconv++, chen2022devil} 
design rotation-invariant geometric features based on relative distances and angles to replace point coordinates as the network input.
While some \cite{kim2020rotation, yu2020deep, li2021closer} propose to use principal component analysis (PCA) to convert coordinates to rotation-invariant canonical poses.
Compared to the hand-crafted geometric features, though PCA operation can preserve more latent information, the ambiguities of the canonical poses need to be alleviated by the extra constraint strategy. Moreover, considering the large amount of LiDAR point clouds, it would be costly to use PCA.
Hence, in our work, we choose to introduce the hand-crafted rotation-invariant geometric features to the feature embedding process by considering the rigorous rotation-invariance property and affordable computation cost.

\section{Method}
\label{sec:method}

\subsection{Overview}
Given a raw LiDAR point cloud $P=\left\{p_{i}\right\}_{i=1}^{N}$ of $N$ points, 3D object detection aims to localize and recognize objects in a 3D scene.
Specifically, each object is represented by a semantic class 
and a 3D oriented bounding box (OBB) parameterized by its center ($c_x$, $c_y$, $c_z$), size ($h$, $w$, $l$) and object orientation $\theta$ from the bird’s-eye view.

\begin{figure}[t]
    \centering
    \includegraphics[width=\linewidth]{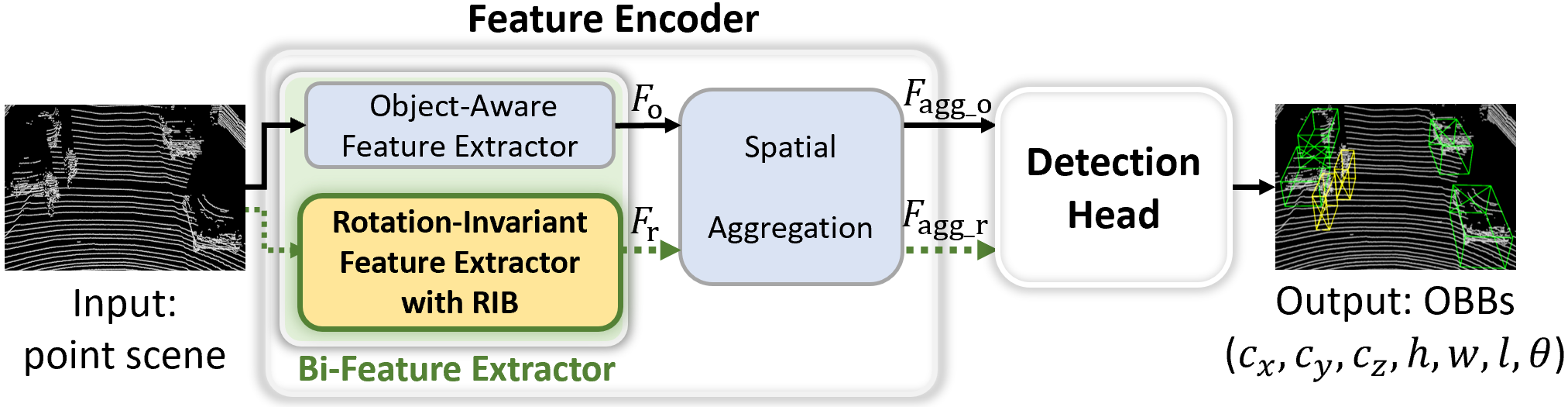}
    \caption{Given the existing point-based detectors consisting of an object-aware feature extractor, a spatial aggregation layer, and a detection head, we further design a bi-feature extractor by incorporating a novel rotation-invariant feature extractor with rotation-invariance block (RIB), thus making existing detectors rotation-robust.
    }
    \label{fig:overview}
\end{figure}

Generally, as shown in Figure \ref{fig:overview}, existing popular point-based detectors are built with two main modules: (1) Feature encoder consisting of an object-aware feature extractor to embed $P$ into object-aware features $F_o$ that encode objects information, and a spatial aggregation layer to further produce local region context feature $F_{agg\_o}$. 
(2) Detection head that recognizes and generates 3D proposals for each foreground object.
Obviously, this design is not rotation-robust since all features are extracted from 3D coordinates, which are rotation-sensitive. 

In this work, we target to \emph{design a plug-and-play modular that can be incorporated into various 3D object detection networks to enhance their rotation robustness}.
To this end, we propose to explicitly leverage object rotation-invariance in 3D detectors, by considering the perfect feature consistency under arbitrary rotation.
As shown in Figure \ref{fig:overview}, the key idea behind our RIDE is to design a bi-feature extractor based on the existing object-aware feature extractor by further introducing a rotation-invariant feature extractor with the rotation-invariance block (RIB) for producing rotation-invariant feature $F_r$ from the input point scene, which is then effectively incorporated into existing spatial aggregation and detection head to 
predict desired rotation-robust OBBs.

Note that, compared to existing rotation-equivariant 3D detectors EON \cite{yu2022rotationally} and TED \cite{wu2023transformation} that achieve rotation robustness via a few discrete pre-defined rotation angles, incorporating rotation-invariance into 3D detectors via our RIDE can ensure robust predictions under \emph{artibrary} and \emph{unseen} rotation angles.
To realize such a flexible and lightweight modular without any restrictions on rotation angle, there are two key challenges in our designs:
(1) How to design and process the rotation-invariant features from large-scale scene points.
(2) How to effectively inject the extracted rotation-invariant features into existing 3D detectors.
Below, we shall present our solutions for the above two challenges.

\subsection{Rotation-Invariant Feature Design}
Given a feature extraction approach $\textit{f}:\mathbb{R} ^{\mathrm{1}\times 3}\rightarrow \mathbb{R} ^{\mathrm{1}\times \mathrm{C}}$, and a point $p_i$ in $P$ with its 3D coordinate ($x$, $y$, $z$). The rotation-invariance can be formulated as:
\begin{equation}
	\label{eq:RI_def}
 f\left( p_i \right) =f\left( p_i R \right) ,
\end{equation}
where $R\in SO\left( 3 \right) $ is a $3 \times 3$ orthogonal rotation matrix. In this case, the features extracted from point cloud $P$ can achieve rotation-invariance to arbitrary $SO(3)$ rotation. 

\begin{figure}[t]
	\centering
	\includegraphics[width=\linewidth]{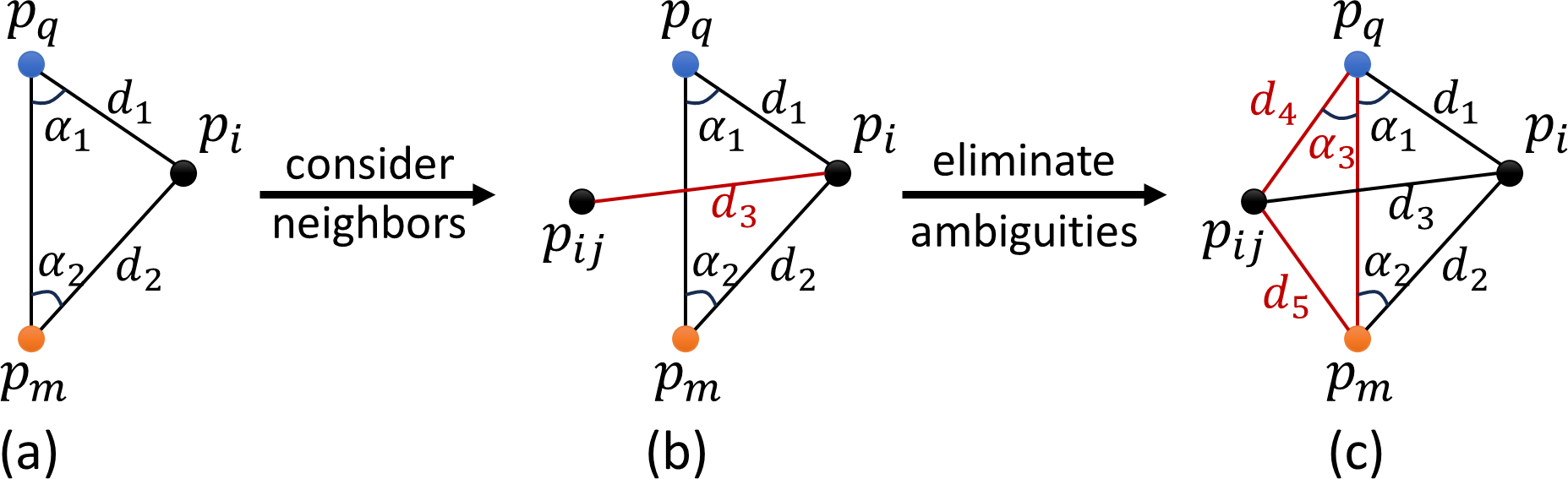}
	\caption{Based on the rotation-invariant feature design (see (a)) proposed in \cite{zhang2019rotation}, (b) we first consider the local structure of a reference point $p_i$ by including its neighbor $p_{ij}$, and then (c) further introduce more extra elements to eliminate ambiguities. 
 Note, $p_m$ and $p_q$ denote the ball center and the geometric barycenter point of $p_i$'s query ball.}
 
	\label{fig:RIF_design}
\end{figure}

Compared to transforming coordinates to rotation-invariant canonical poses, extracting relative geometric features based on relative distances and angles is more computationally stable and efficient. 
Since existing point-based detectors exploit local regions within the query ball, an early work \cite{zhang2019rotation} constructs a triangle structure consisting of a reference point $p_i$ in the query ball,
its ball center point $p_m$ and the geometric barycenter point $p_q$; see Figure \ref{fig:RIF_design}(a).
Then the rotation-invariant feature (RIF) of $p_i$ is: 
\begin{equation}
	\label{eq:RIF_riconv}
 \begin{array}{c}
 \text{RIF}(p_i)=\left\{ \left\| d_1 \right\| _2, \left\| d_2 \right\| _2, \cos(\alpha _1), \cos(\alpha _2) \right\},\\ 
\end{array}
\end{equation}
where $\left\| d_1 \right\| _2$ and $\left\| d_2 \right\| _2$ is the distance from $p_i$ to $p_q$ and $p_m$, respectively. The angle $\alpha_1$ and $\alpha_2$ subtends at $p_q$ and $p_m$, respectively.
However, such a simple RIF design ignores the exploration of local neighbors of $p_i$, thus losing much geometric information compared with the coordinates input.



To effectively exploit the local point-pair relations, we further consider the geometric structure between $p_i$ and its adjacent neighbors.
As Figure \ref{fig:RIF_design}(b) shows, given $p_i$ and one of its neighbors $p_{ij}$ in a query ball, we further introduce $d_3$ to measure the geometric relation between $p_i$ and $p_{ij}$:
\begin{equation}
	\label{eq:RIF_riconv_refine}
 \text{RIF}(p_i,p_{ij})= \left\{ \left\| d_1 \right\| _2, \left\| d_2 \right\| _2, \left\| d_3 \right\| _2, \cos(\alpha _1), \cos(\alpha _2) \right\}.
\end{equation}

However, introducing $p_{ij}$ in such a way brings ambiguity to describe $p_i$'s local neighborhood.
That is, relying only on the five elements in Eq.~\eqref{eq:RIF_riconv_refine} cannot uniquely locate $p_{ij}$ relative to $p_i$. 
As long as $p_{ij}$ is located on a circle with $p_i$ as the center and $\left\| d_3 \right\| _2$ as the radius, the RIF is always the same.


To eliminate this ambiguity, as shown in Figure~\ref{fig:RIF_design}(c), we further extend Eq.~\eqref{eq:RIF_riconv_refine} by considering the relative distances and angles brought by the triangle $p_q$-$p_m$-$p_{ij}$, i.e., the distance $\left\| d_4 \right\| _2$ from $p_{ij}$ to $p_q$, the distance $\left\| d_5 \right\| _2$ from $p_{ij}$ to $p_m$, as well as the angle $\alpha_3$ subtended at $p_q$. 
One can simply verify that the location of $p_{ij}$ can be identified uniquely using the eight elements in Figure~\ref{fig:RIF_design}(c); see the proof in the supplementary material.

Unlike a single object that can rotate around any axis, a 3D scene can only rotate around the gravity axis, i.e., the $z$-axis, during scanning.
In other words, the $z$ coordinates of the LiDAR-based points will not change with rotation.
With this prior, 
the $SO(3)$ rotation degenerates to the $SO(2)$ rotation around $z$-axis. 
To retain as much information as possible, we finally re-formulate Eq.~\eqref{eq:RIF_riconv_refine} as follows:
\begin{equation}
	\label{eq:2drif}
 \begin{aligned}
\text{RIF}_{\text{2D}}(p_i,p_{ij})=\{ 
	&\left\| d_1 \right\| _2, \left\| d_2 \right\| _2, \left\| d_3 \right\| _2, \left\| d_4 \right\| _2, \left\| d_5 \right\| _2, \\
	&\cos \left( \alpha _1 \right), \cos \left( \alpha _2 \right), \cos \left( \alpha _3 \right),
 z_i \}.
 \end{aligned}
\end{equation}
Note that, the calculation of distances and angles in Eq.~\eqref{eq:2drif} is conducted via 2D coordinates ($x$, $y$), thus quite efficient.

\paragraph{Rotation-invariance block.}
In summary, given one query ball in $m$ type that contains $N_s$ points, we first use Eq.~\eqref{eq:2drif} to extract rotation-invariant features of each point with the dimension $N_s \times 9$, where we only take one adjacent neighbor point in the clockwise direction as $p_{ij}$ for efficiency.
Next, we employ a series of MLPs and max pooling to obtain the rotation-invariant features with the dimensions ${1 \times C}$ for one query ball. Assuming a point cloud $P$ includes $N$ points, thus we have $N$ query balls in $m$ type and acquire the feature representation $F_{rm} \in \mathbb{R} ^{\mathrm{N}\times \mathrm{C}}$, which we term the module of the above processing as rotation-invariance block (RIB).  


\subsection{Network Architecture}
In order to leverage RIB for detection, we propose a bi-feature extractor and a corresponding detection head. In this way, we can equip the existing detectors with our RIDE.

\begin{figure}[t]
	\centering
	\includegraphics[width=\linewidth]{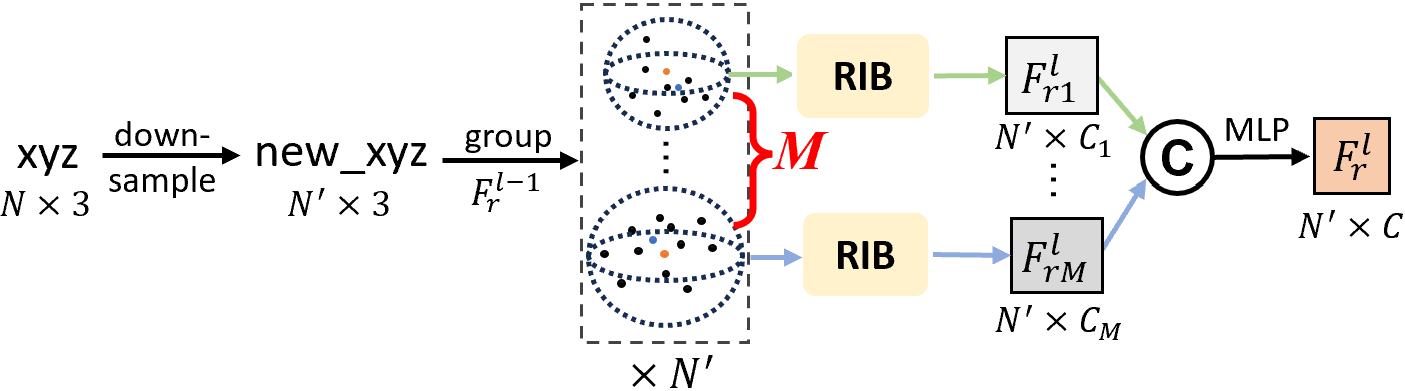}
	\caption{The detail of extracting RIF representation $F_r^l$ at $l$-th bi-set abstraction (Bi-SA) layer. Meanwhile, the OAF representation $F_o^l$ can be acquired by replacing RIB with MM (MLPs + Max pooling) module. Therefore, we can parallel RIBs and MMs to obtain both $F_r$ and $F_o$ at one layer.}
	\label{fig:bi-sa}
\end{figure}

\paragraph{Bi-feature extractor.}
To obtain $F_o$ and $F_r$ from the feature extractors in Figure \ref{fig:overview}, we incorporate these two feature extractors and propose the bi-feature extractor consisting of $L$ bi-set abstraction (Bi-SA) layers. We show the detail of extracting $F_r^l$ at $l$-th Bi-SA layer in Figure \ref{fig:bi-sa}. Specifically, given an input point cloud with $N$ points, we first down-sample it to $N'$ points and group them with the $(l-1)$-th layer feature representations $F_r^{l-1}$ via query balls. Suppose we have $M$ types of query balls (with different radii, number of sample points, etc.), we utilize the proposed RIBs to obtain $F_{rm}^{l}$, where $m$ denotes one type of the query balls. Finally, we concatenate all $M$ types of representations and produce the output $F_r^l$ of the $l$-th Bi-SA layer. Next, we can just replace RIB with the MM (MLPs + Max pooling) module, and replace $F_r^{l-1}$ with $F_o^{l-1}$ to extract $F_o^l$. In this way, we obtain both rotation-invariant features representation $F_r^l$ and object-aware features representation $F_o^l$ simultaneously by paralleling RIBs and MMs at one layer. Being processed by $L$ Bi-SA layers, we can get the desired $F_o$ and $F_r$ with the bi-feature extractor.

\paragraph{Detection head.}
After acquiring aggregated feature $F_{agg\_o}$ and $F_{agg\_r}$ from $F_o$ and $F_r$ via spatial aggregation layer, respectively, we propose to use $F_{agg\_o}$ to decode orientation $\theta$ of oriented bounding boxes (OBBs), since $\theta$ is the rotation-sensitive attribute, and use $F_{agg\_r}$ to decode the semantic-related attributes of the OBBs and strengthen the rotation robustness. Particularly, We divide the attributes of OBB into three parts: (I) the category of OBB, (II) the orientation of OBB, and (III) the rest (i.e., the center point coordinate and the size of OBB). On the other hand, we propose three types of input to the detection head: (a) $F_{agg\_r}$, (b) $F_{agg\_o}$, and (c) $F_{agg\_m}$, i.e., the fusion of $F_{agg\_r}$ and $F_{agg\_o}$ (concatenation by default). The specific decode pairing between input types (a-c) and OBB attributes (I-III) will be elaborated in the Experiments Section.

\section{Experiments}
\label{sec:exp}

\begin{table*}[t]
	\renewcommand{\arraystretch}{0.75}
    \setlength{\tabcolsep}{1mm}
	\begin{center}
			\begin{tabular}{ c | l | c | c c c | c c c | c c c | c | c }
                \toprule[1pt]
                \multirow{2}{*}{Type} & \multirow{2}{*}{Method} & \multirow{2}{*}{Case} & \multicolumn{3}{c |}{Car (IoU=0.7)} & \multicolumn{3}{c |}{Pedestrian (IoU=0.5)} & \multicolumn{3}{c |}{Cyclist (IoU=0.5)} &  \multirow{2}{*}{$\mathbf{\mathbf{\Delta} }$} & \multirow{2}{*}{FPS} \\
                ~ & ~ & ~ & Easy & Mod. & Hard & Easy & Mod. & Hard & Easy & Mod. & Hard & ~ & ~ \\
                \hline
                \multirow{8}{*}{1-stage} & \multirow{2}{*}{3DSSD \shortcite{yang20203dssd}} & \textit{DR} & 89.2 & 80.3 & 77.2 & 59.6 & 52.8 & 47.7 & 91.2 & 71.1 & 66.8 & \multirow{2}{*}{123.6} & \multirow{2}{*}{63.1} \\
                ~ & ~ & \textit{AR} & 71.1 & 58.5 & 54.3 & 50.7 & 46.0 & 41.4 & 78.8 & 57.0 & 54.5 & ~ & ~ \\
                \cline{2-14}
                ~ & \multirow{2}{*}{RIDE-3DSSD (Ours)} & \cellcolor{gray!10}\textit{DR} & \cellcolor{gray!10}\textbf{92.2} & \cellcolor{gray!10}\textbf{82.4} & \cellcolor{gray!10}\textbf{79.6} & \cellcolor{gray!10}58.7 & \cellcolor{gray!10}51.8 & \cellcolor{gray!10}47.1 & \cellcolor{gray!10}\textbf{92.3} & \cellcolor{gray!10}\textbf{74.0} & \cellcolor{gray!10}\textbf{69.5} & \multirow{2}{*}{\textbf{85.4}} & \multirow{2}{*}{43.7} \\
                ~ & ~ & \cellcolor{gray!10}\textit{AR} & \cellcolor{gray!10}\textbf{75.8} & \cellcolor{gray!10}\textbf{63.2} & \cellcolor{gray!10}\textbf{58.6} & \cellcolor{gray!10}\textbf{55.8} & \cellcolor{gray!10}\textbf{49.9} & \cellcolor{gray!10}\textbf{45.5} & \cellcolor{gray!10}\textbf{87.2} & \cellcolor{gray!10}\textbf{65.1} & \cellcolor{gray!10}\textbf{61.1} & ~ & ~ \\
                \cline{2-14}
                ~ & \multirow{2}{*}{IA-SSD \shortcite{zhang2022not}} & \textit{DR} & 90.7 & 80.3 & 77.2 & 57.4 & 52.4& 47.3 & 85.1 & 68.6 & 65.4 & \multirow{2}{*}{110.1} & \multirow{2}{*}{70.2} \\
                ~ & ~ & \textit{AR} & 73.9 & 62.2 & 57.9 & 49.1 & 44.2 & 39.9 & 74.6 & 57.8 & 54.7 & ~ & ~ \\
                \cline{2-14}
                ~ & \multirow{2}{*}{RIDE-IA-SSD (Ours)} & \cellcolor{gray!10}\textit{DR} & \cellcolor{gray!10}88.8 & \cellcolor{gray!10}\textbf{81.6} & \cellcolor{gray!10}\textbf{77.4} & \cellcolor{gray!10}53.7 & \cellcolor{gray!10}48.0 & \cellcolor{gray!10}43.9 & \cellcolor{gray!10}\textbf{86.7} & \cellcolor{gray!10}65.9 & \cellcolor{gray!10}62.2 & \multirow{2}{*}{\textbf{52.1}} & \multirow{2}{*}{62.7} \\
                ~ & ~ & \cellcolor{gray!10}\textit{AR} & \cellcolor{gray!10}\textbf{81.2} & \cellcolor{gray!10}\textbf{68.2} & \cellcolor{gray!10}\textbf{63.3} & \cellcolor{gray!10}\textbf{52.8} & \cellcolor{gray!10}\textbf{47.9} & \cellcolor{gray!10}\textbf{44.0} & \cellcolor{gray!10}\textbf{80.1} & \cellcolor{gray!10}\textbf{60.9} & \cellcolor{gray!10}\textbf{57.9} & ~ & ~ \\
                \hline
                \multirow{4}{*}{2-stage} & \multirow{2}{*}{PointRCNN \shortcite{shi2019pointrcnn}} & \textit{DR} & 91.5 & 80.3 & 77.8 & 63.2 & 55.5 & 48.5 & 88.5 & 71.3 & 66.5 & \multirow{2}{*}{60.1} & \multirow{2}{*}{25.4} \\ 
                ~ & ~ & \textit{AR} & 84.1 & 67.5 & 64.9 & 56.5 & 48.5 & 43.1 & 87.2 & 67.9 & 63.3 & ~ & ~ \\ 
                \cline{2-14}
                ~& \multirow{2}{*}{RIDE-PointRCNN (Ours)} & \cellcolor{gray!10}\textit{DR} & \cellcolor{gray!10}89.5 & \cellcolor{gray!10}80.0 & \cellcolor{gray!10}77.6 & \cellcolor{gray!10}60.1 & \cellcolor{gray!10}52.8 & \cellcolor{gray!10}46.1 & \cellcolor{gray!10}\textbf{90.2} & \cellcolor{gray!10}69.5 & \cellcolor{gray!10}65.0 & \multirow{2}{*}{\textbf{56.6}} & \multirow{2}{*}{22.9} \\ 
                ~ & ~ & \cellcolor{gray!10}\textit{AR} & \cellcolor{gray!10}\textbf{85.8} & \cellcolor{gray!10}\textbf{70.6} & \cellcolor{gray!10}63.9 & \cellcolor{gray!10}\textbf{57.2} & \cellcolor{gray!10}\textbf{50.7} & \cellcolor{gray!10}\textbf{44.1} & \cellcolor{gray!10}83.4 & \cellcolor{gray!10}61.5 & \cellcolor{gray!10}57.0 & ~ & ~ \\ 
                \bottomrule[1pt]
		  \end{tabular}
	\end{center}
    \caption{The results of 3D object detection on the KITTI \textit{val} set. We report the average precision (\%) over 40 recall positions (AP\_R40) and the FPS under default rotation (\textit{DR}) and arbitrary $SO(2)$ rotation (\textit{AR}) cases, and the rotation robustness via the difference $\mathbf{\mathbf{\Delta} }$. IoU threshold of AP\_R40 is 0.7 for Car and 0.5 for Pedestrian / Cyclist, respectively.}
	\label{tab:kitti_val}
\end{table*}

\subsection{Datasets and Evaluation Metrics}

\paragraph{Datasets.} To evaluate the object detection performance, we employ two popular benchmarks: KITTI \cite{geiger2012we} and nuScenes \cite{caesar2020nuscenes}.
\paragraph{KITTI} contains 7481 training samples 
for 3D object detection in autonomous driving scenario. 
We 
follow the recent works \cite{chen2022sasa, wu2023transformation} to divide it into \textit{training} and \textit{val} sets with 3712 and 3769 samples, respectively.

\paragraph{nuScenes} provides over 390k LiDAR sweeps in 1000 scenes, which is a challenge for autonomous driving scenario due to its size and complexity. The dataset includes 1.4 million 3D objects across 10 categories, along with object attributes and velocity, with approximately 40000 points per frame. 

\paragraph{Evaluation metrics.} 
The official evaluation metric for 3D object detection is the 3D average precision (AP) (\%) with 40 recall positions.
To assess the rotation robustness, we test the AP performance in
(1) the point clouds rotated with random angles from $\left[ -\pi /4,\pi /4 \right] $ (i.e. the default setting, denoted as \textit{DR}), and (2) the point clouds rotated with random angles from $\left[ -\pi ,\pi \right]$ (i.e. the arbitrary $SO(2)$ rotation along $z$-axis, denoted as \textit{AR}), then we calculate the difference $\mathbf{\mathbf{\Delta}}$ between the two.
Formally,
\begin{equation}
	\label{eq:robustness}
\mathbf{\mathbf{\Delta}}=\left| \varPhi(\mathrm{AP}_{DR} - \mathrm{AP}_{AR}) \right|,
\end{equation}
where $\varPhi(\cdot)$ denotes the sum operation of the AP across all categories.
Obviously, a higher AP under both \textit{DR} and \textit{AR} cases and a smaller $\mathbf{\mathbf{\Delta}}$ represent a more robust model. 
For the nuScenes, we report the nuScenes detection score (NDS) as the extra metric for overall performance. Particularly, NDS comprises mean AP (mAP) and 5 mean average errors of translation, scale, orientation, velocity, and attribute:
\begin{equation}
	\label{eq:nds}
\mathrm{NDS}=\frac{1}{10}\left[ 5\mathrm{mAP}+\sum_{\mathrm{mTP}\in \mathrm{TP}}{\left( 1-\min \left( 1, \mathrm{mTP} \right) \right)} \right],
\end{equation}
where $\mathrm{TP}$ denotes the set of the 5 mean average errors.

\begin{table}[t]
	\renewcommand{\arraystretch}{1.0}
    \setlength{\tabcolsep}{1mm}
	\begin{center}
			\begin{tabular}{l | c c c | c }
                \toprule[1pt]
                \multirow{2}{*}{Method} & \multicolumn{3}{c |}{Car (IoU=0.7)} & \multirow{2}{*}{FPS} \\
                ~ & Easy & Moderate & Hard &  ~ \\
                \hline
                EON-PointRCNN \shortcite{zhao2022rotation} & 89.1 & 78.6 & \underline{77.6} & 2.5 \\
                TED-S \shortcite{wu2023transformation} & \textbf{93.1} & \textbf{87.9} & \textbf{85.8} & 11.1 \\
                \hline
                RIDE-PointRCNN (Ours) & \underline{89.5} & \underline{80.0} & \underline{77.6} & \textbf{22.8} \\
                \bottomrule[1pt]
		  \end{tabular}
	\end{center}
    \caption{The comparison with the related works EON \cite{yu2022rotationally} and TED \cite{wu2023transformation} that both utilize two-stage frameworks. We show the APs in Car category and the FPS, all of which are conducted under the \textit{DR} case.}
	\label{tab:eon_ted}
\end{table}

\begin{figure*}[t]
	\centering
	\includegraphics[width=.9\linewidth]{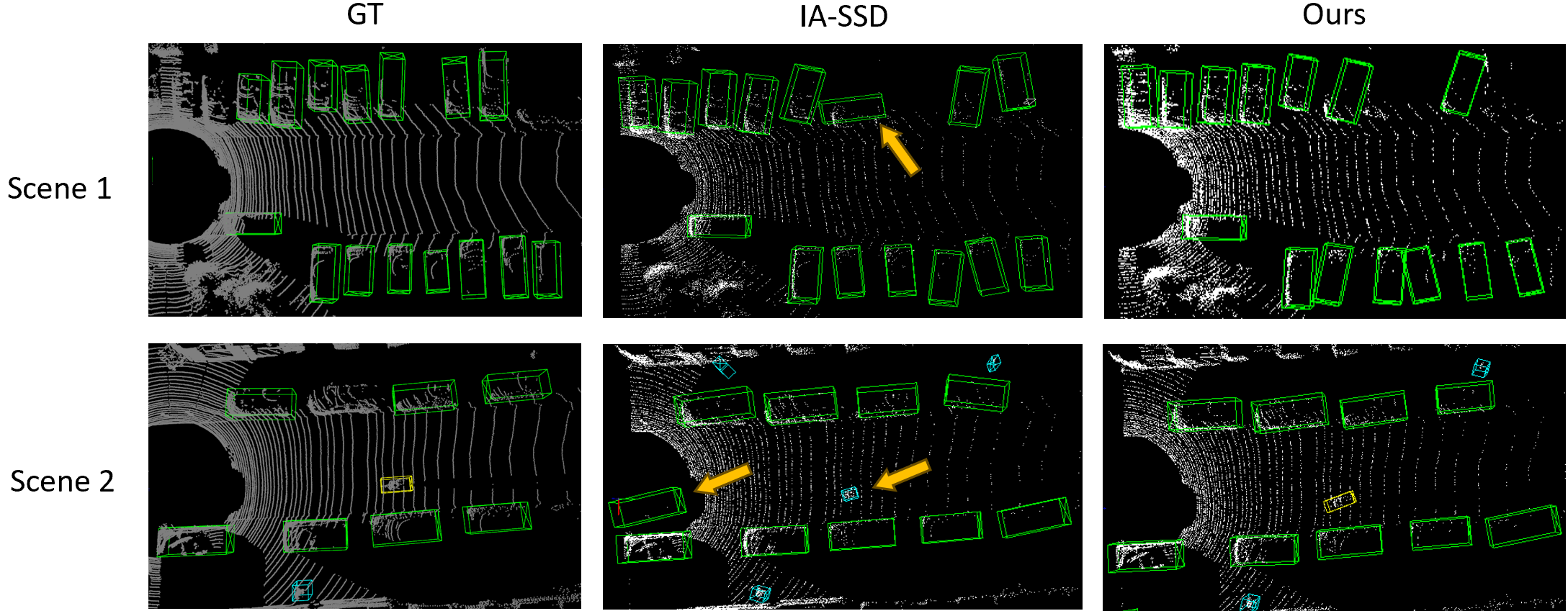}
	\caption{Qualitative comparisons between our RIDE-IA-SSD with IA-SSD \cite{zhang2022not} against the ground truth (GT). The yellow arrows denote the distinct differences.}
	\label{fig:vis}
\end{figure*}

\subsection{Implementation Details}
We choose all the point-based detectors to our best knowledge as the baselines to evaluate the performance and compatibility of our method, namely 3DSSD \cite{yang20203dssd} and IA-SSD \cite{zhang2022not} for the one-stage detectors and PointRCNN \cite{shi2019pointrcnn} for the two-stage detector, all of which are supported by the OpenPCDet \cite{openpcdet2020} toolbox. 


\paragraph{One-stage detectors.} 
To equip existing one-stage detectors (i.e., 3DSSD and IA-SSD) with our RIDE, we keep the improved down-sampling strategies on each detector baseline and replace the original feature extractor in the encoder with our bi-feature extractor. 
On the KITTI, we train all the models using a single NVIDIA GeForce RTX 3090 GPU, with ADAM optimizer for 80 epochs, in which the initial learning rate is set to 0.01 with one cycle scheduler \cite{smith2019super} and follow the default data augmentation strategy.
On the nuScenes, we train the original 3DSSD and our modified version (denoted as RIDE-3DSSD) using four NVIDIA GeForce RTX 3090 GPUs, with ADAM optimizer for 20 epochs, in which the initial learning rate is set to 0.002.
We train the original IA-SSD and ours (denoted as RIDE-IA-SSD) using three 3090 GPUs with an initial learning rate of 0.01.
Empirically, we use the decode pairing (c-I, b-II, c-III) for one-stage detectors by default.


\paragraph{Two-stage detector.}  As PointRCNN has two stages for refined proposals, we only modify its first-stage pipeline, which employs the encoder of PointNet++ \cite{qi2017pointnet++} as the backbone. We replace the set abstraction (SA) layers with our Bi-SA layers, while changing the feature propagation (FP) layers accordingly. The models are trained on the KITTI with a batch size of 4 on a single NVIDIA GeForce RTX 3090 GPU, using ADAM optimizer for 80 epochs with an initial learning rate of 0.01, a one cycle scheduler, and default data augmentation strategy. Empirically, we use the decode pairing (c-I, b-II, c-III) for the first-stage decoder.

\begin{table*}[t]
	\renewcommand{\arraystretch}{0.8}
    \setlength{\tabcolsep}{1mm}
	\begin{center}
			\begin{tabular}{ l | c | c c c c c c c c c c | c | c }
                \toprule[1pt]
                Method & Case & Car & Truck & Bus & Trailer & C.V. & Ped. & Moto. & Bicycle & T.C. & Barrier & $\mathbf{\Delta}$ & NDS \\
                \hline
                \multirow{2}{*}{3DSSD \shortcite{yang20203dssd}} & \textit{DR} & 74.1 & 37.2 & 60.5 & 34.6 & 14.8 & 55.4 & 28.7 & 11.2 & 14.4 & 34.4 & \multirow{2}{*}{26.9} & 55.9 \\
                ~ & \textit{AR} & 71.6 & 36.9 & 59.3 & 29.2 & 13.5 & 55.2 & 24.7 & 8.8 & 11.2 & 28.0 & ~ & 46.6 \\
                \cline{1-14}
                \multirow{2}{*}{RIDE-3DSSD (Ours)} & \cellcolor{gray!10}\textit{DR} & \cellcolor{gray!10}71.6 & \cellcolor{gray!10}\textbf{44.3} & \cellcolor{gray!10}\textbf{68.6} & \cellcolor{gray!10}\textbf{37.5} & \cellcolor{gray!10}\textbf{17.0} & \cellcolor{gray!10}52.8 & \cellcolor{gray!10}\textbf{33.3} & \cellcolor{gray!10}\textbf{14.3} & \cellcolor{gray!10}13.6 & \cellcolor{gray!10}23.5 & \multirow{2}{*}{\textbf{21.6}} & 56.3 \\ 
                ~ & \cellcolor{gray!10}\textit{AR} & \cellcolor{gray!10}69.3 & \cellcolor{gray!10}\textbf{43.7} & \cellcolor{gray!10}\textbf{66.5} & \cellcolor{gray!10}\textbf{30.6} & \cellcolor{gray!10}\textbf{16.3} & \cellcolor{gray!10}51.8 & \cellcolor{gray!10}\textbf{32.3} & \cellcolor{gray!10}\textbf{12.3} & \cellcolor{gray!10}\textbf{12.3} & \cellcolor{gray!10}19.8 & ~ & 47.1 \\ 
                \cline{1-14}
                \multirow{2}{*}{IA-SSD \shortcite{zhang2022not}} & \textit{DR} & 73.6 & 45.1 & 67.2 & 31.5 & 16.1 & 66.3 & 36.6 & 15.3 & 28.4 & 54.8 & \multirow{2}{*}{31.6} & 48.6 \\
                ~ & \textit{AR} & 69.6 & 43.1 & 65.9 & 25.9 & 15.3 & 65.1 & 33.4 & 11.7 & 25.3 & 48.0 & ~ & 46.5 \\
                \cline{1-14}
                \multirow{2}{*}{RIDE-IA-SSD (Ours)} & \cellcolor{gray!10}\textit{DR} & \cellcolor{gray!10}\textbf{74.7} & \cellcolor{gray!10}\textbf{47.9} & \cellcolor{gray!10}\textbf{68.7} & \cellcolor{gray!10}31.4 & \cellcolor{gray!10}\textbf{20.9} & \cellcolor{gray!10}\textbf{71.3} & \cellcolor{gray!10}\textbf{45.2} & \cellcolor{gray!10}\textbf{21.9} & \cellcolor{gray!10}\textbf{39.3} & \cellcolor{gray!10}\textbf{55.9} & \multirow{2}{*}{\textbf{23.6}} & 50.8 \\ 
                ~ & \cellcolor{gray!10}\textit{AR} & \cellcolor{gray!10}\textbf{72.5} & \cellcolor{gray!10}\textbf{46.3} & \cellcolor{gray!10}\textbf{66.9} & \cellcolor{gray!10}\textbf{28.0} & \cellcolor{gray!10}\textbf{19.2} & \cellcolor{gray!10}\textbf{70.0} & \cellcolor{gray!10}\textbf{42.3} & \cellcolor{gray!10}\textbf{16.7} & \cellcolor{gray!10}\textbf{38.3} & \cellcolor{gray!10}\textbf{53.4} & ~ & 49.0 \\ 
                \bottomrule[1pt]
		  \end{tabular}
	\end{center}
    \caption{The results of 3D object detection on the nuScenes \textit{val} set. We report the evaluation metrics including APs (\%) in 10 categories, NDS (\%) under default rotation (\textit{DR}) and arbitrary $SO(2)$ rotation (\textit{AR}) cases
    , and the $\mathbf{\mathbf{\Delta} }$ between the two cases. C.V., Ped., Moto. and T.C. are short for Construction Vehicle, Pedestrian, Motorcycle and Traffic Cone categories, respectively. 
    }
	\label{tab:nus_val}
\end{table*}

\subsection{3D Object Detection on KITTI}


We first evaluate each model and its version equipped with our RIDE on the \textit{val} set of KITTI \cite{geiger2012we}; see Table \ref{tab:kitti_val} for the results.
Clearly, under the \textit{AR} case 
, when equipping with our method, the two one-stage methods achieve higher AP values than the original methods in all categories, with the increment of 5.6\% and 4.6\% in mean AP (mAP), respectively. Even for the two-stage method with refinement, our method achieves higher AP values in most categories.
Next, under the \textit{DR} case 
, our RIDE is even beneficial to the original methods in Car and Cyclist categories as well, and a gain of 1.3\% of mAP for the RIDE-3DSSD.
It is worth noting that when equipped with our RIDE, the predictions across all categories of all methods are no longer that sensitive to rotation, i.e., the difference $\mathbf{\Delta}$ values become significantly smaller, and the rotation robustness is boosted by 31\%, 53\%, and 6\%, respectively. 

From the results in Table \ref{tab:kitti_val}, it can be observed our method can: (1) significantly boost the APs under the \textit{AR} case across all categories, and (2) improve the APs under the \textit{DR} case for certain categories. While boosting the precision performance, the inference speed of our models (see the last column of Table \ref{tab:kitti_val}) testing on a single 3090 GPU still achieves a competitive result, with decreases of 31\%, 12\%, and 10\% compared to the baselines, respectively. 

We demonstrate some qualitative comparisons in Figure \ref{fig:vis} (see more in the supplementary material), which compares our RIDE-IA-SSD with its baseline against the ground truth (GT) in $+ \pi /2$ rotation angle.  
It is clear that the predicted oriented bounding boxes (OBBs) of our method achieve more precise results compared with IA-SSD \cite{zhang2022not}, whereas the results of IA-SSD show some obvious wrong predictions; see the OBBs marked by yellow arrows.
Intuitively, our RIDE, contributing to the rotation-invariance representation of the semantic information of the OBB, is the key factor for the improvement, which is supported by the orientation of the predicted OBBs in Figure \ref{fig:vis}.

We further compare our model with two existing rotation-robust methods EON \cite{yu2022rotationally} and TED \cite{wu2023transformation}; see Table \ref{tab:eon_ted} for the results. Note that both EON and TED lack feasible code to test under the arbitrary rotation (\textit{AR}) case, so we use the results reported in their paper, which are in \textit{DR} case. Compared with the EON-PointRCNN, which also incorporates proposed modules into the baseline, the PointRCNN \cite{shi2019pointrcnn} integrated with our RIDE achieves better results with a much higher inference speed. On the other hand, compared to the voxel-based TED-S, although the AP values of our model are inferior, our inference speed is faster.

\subsection{3D Object Detection on nuScenes}

The evaluation on the \textit{val} set of nuScenes \cite{caesar2020nuscenes} is reported in Table \ref{tab:nus_val}. Our models are notably better than the corresponding baselines under both \textit{DR} and \textit{AR} cases. Specifically, our method improves +1.2\% and +1.7\% mAP under the \textit{DR} and \textit{AR} cases using RIDE-3DSSD, respectively, with a great advantage in Truck, Bus and Motorcycle categories. 
Moreover, we achieve better APs across all categories using RIDE-IA-SSD, and lead to a +4.2\% and +5.1\% mAP improvement under the \textit{DR} and \textit{AR} cases, respectively. The rotation robustness boosts 19\% and 28\% compared with the baselines, respectively. The details of NDS can be found in the supplementary material.

Consistent with the KITTI, the results on the nuScenes also demonstrate significant improvement due to the introduction of our method. With the contribution of the rotation-invariance representation, the baselines equipped with our RIDE achieve better results in almost all categories. We observe that the difference $\mathbf{\mathbf{\Delta} }$ boost is not as much as on the KITTI. We assume the reason comes from the complexity and large scale of the dataset, as the nuScenes is a challenging dataset for object detection. 

\begin{table}[t]
	\renewcommand{\arraystretch}{1.0}
    \setlength{\tabcolsep}{1mm}
	\begin{center}
			\begin{tabular}{ l | c c c | c }
                \toprule[1pt]
                RIF method & Easy & Moderate & Hard & mAP \\
                \hline
                Baseline \cite{yang20203dssd} & 71.1 & 58.5 & 54.3 & 61.3 \\
                $\text{RIF}(p_i)$ & 73.3 & 59.8 & 56.7 & 63.3 \\
                $\text{RIF}(p_i,p_{ij})$ & \underline{73.8} & \underline{61.3} & \underline{57.1} & \underline{64.1} \\
                \hline
                $\text{RIF}_{\text{2D}}(p_i,p_{ij})$ & \textbf{75.8} & \textbf{63.2} & \textbf{58.6} & \textbf{65.9} \\
                \bottomrule[1pt]
		  \end{tabular}
	\end{center}
    \caption{The comparison between proposed $\text{RIF}_{\text{2D}}$ performance under the arbitrary rotation (\textit{AR}) case with the referred RIF methods and baseline. The $\text{RIF}(p_i)$, $\text{RIF}(p_i,p_{ij})$ and $\text{RIF}_{\text{2D}}(p_i,p_{ij})$ denote for the rotation-invariant features design by Eq.~\eqref{eq:RIF_riconv},~\eqref{eq:RIF_riconv_refine} and~\eqref{eq:2drif}, respectively.}
	\label{tab:ablation_rif}
\end{table}

\begin{table}[t]
	\renewcommand{\arraystretch}{1.0}
    \setlength{\tabcolsep}{1mm}
	\begin{center}
			\begin{tabular}{ l | c c c | c }
                \toprule[1pt]
                Fusion strategy & Easy & Moderate & Hard & mAP \\
                \hline
                SK-Conv \cite{li2019selective} & 74.4 & 62.3 & 58.2 & 65.0 \\
                Concatenation (Ours) & 75.8 & 63.2 & 58.6 & 65.9 \\
                \bottomrule[1pt]
		  \end{tabular}
	\end{center}
    \caption{The comparison of feature fusion strategies between concatenation (proposed) and the SK-Conv \cite{li2019selective}. }
	\label{tab:ablation_fusion}
\end{table}

\begin{table}[t]
	\renewcommand{\arraystretch}{1.0}
    \setlength{\tabcolsep}{1mm}
	\begin{center}
			\begin{tabular}{ c c c | c c c | c }
                \toprule[1pt]
                I & II & III & Easy & Moderate & Hard & mAP \\
                \hline
                $F_{agg\_r}$ & $F_{agg\_o}$ & $F_{agg\_r}$ & 75.7 & 60.6 & 55.3 & 63.9 \\
                $F_{agg\_r}$ & $F_{agg\_o}$ & $F_{agg\_o}$ & 74.9 & 60.6 & 55.9 & 63.8 \\
                $F_{agg\_r}$ & $F_{agg\_o}$ & $F_{agg\_m}$ & 70.2 & 58.7 & 54.4 & 61.1 \\
                $F_{agg\_m}$ & $F_{agg\_o}$ & $F_{agg\_r}$ & 74.5 & 60.5 & 55.8 & 63.6 \\
                $F_{agg\_m}$ & $F_{agg\_o}$ & $F_{agg\_o}$ & \underline{75.8} & \textbf{63.2} & \textbf{58.6} & \textbf{65.9} \\
                $F_{agg\_m}$ & $F_{agg\_o}$ & $F_{agg\_m}$ & \textbf{76.0} & \underline{62.4} & \underline{57.7} & \underline{65.4} \\
                \bottomrule[1pt]
		  \end{tabular}
	\end{center}
    \caption{The comparison of decode pairings between our proposed (bottom row) and the other potential pairing combinations. 
    The first three columns represent the three attribute parts of OBB: (I) category, (II) orientation, and (III) center coordinate and box size, where $F_{agg\_r}$, $F_{agg\_o}$ and $F_{agg\_m}$ refer to a, b and c in the decode pairing, respectively.}
	\label{tab:ablation_pairings}
\end{table}

\subsection{Ablation Study}
We conduct ablation studies to verify the effectiveness of each part of our method, all of which are evaluated using RIDE-3DSSD in Car category on the KITTI \textit{val} set under the arbitrary rotation (\textit{AR}) case. 

\paragraph{Effect of rotation-invariant features.}
To validate the design of the $\text{RIF}_{\text{2D}}$ proposed in Eq.~\eqref{eq:2drif}, we replace it with the rotation-invariant features (RIFs) in Eq.~\eqref{eq:RIF_riconv} and~\eqref{eq:RIF_riconv_refine} and conduct experiments, respectively. Results are shown in Table \ref{tab:ablation_rif}, with the $\text{RIF}_{\text{2D}}$ design significantly outperforming the baseline by 4.6\%. On the other hand, the performance of the $\text{RIF}(p_i,p_{ij})$ shows the introduction of the neighbors $p_{ij}$ leads to a +0.8\% mAP improvement, indicating a notable contribution of the local geometric structure.

\paragraph{Effect of feature fusion strategy.}

Next, we compare the fusion approaches for obtaining $F_{agg\_m}$ from fusing aggregated feature representations $F_{agg\_r}$ and $F_{agg\_o}$: (1) concatenation and (2) attention-based method. Intuitively, the attention mechanism has been demonstrated as an effective way to enhance the fusion feature representation, and we use the SK-Conv in SKNet \cite{li2019selective} as the attention-based fusion module for the evaluation. However, the results in Table \ref{tab:ablation_fusion} illustrate the performance of the attention-based method is inferior to the direct concatenation one. We assume it is caused by the distinct properties of rotation-invariance and object-awareness, which are undermined due to the re-weighting feature operation from attention.

\paragraph{Effect of decode pairing.}
In our method, the decode pairing is a variable that needs to be pre-determined. According to the property of rotation-invariance, we only consider introducing $F_{agg\_r}$ to decode the semantic information of the OBBs. To this end, we explore the different combinations of decode pairing for one-stage detector as shown in Table \ref{tab:ablation_pairings}. The results demonstrate the decode pairing of (c-I, b-II, b-III) achieves the best performance, thus we use it as the default pairing setting.

\section{Conclusion}
\label{sec:conclusion}
In this paper, we present RIDE, a plug-and-played 3D point-based detector modular incorporating rotation-invariance feature representation. To our best knowledge, RIDE is the first attempt to exploit the rotation-invariant features within local geometric structure for the 3D object detection task. Specifically, we propose a bi-feature extractor with bi-set abstraction (Bi-SA) layers to extract both rotation-invariant and object-aware features, and decode the attributes of the oriented bounding boxes (OBBs) according to the characteristics of the features. RIDE can be easily integrated into the existing state-of-the-art one-stage and two-stage detectors, and extensive experiments on the benchmarks showcase that our method can significantly improve both performance and rotation robustness simultaneously. 

\paragraph{Limitations.} Our RIDE has some inevitable limitations. (1) Although the proposed design of RIFs can represent a relative local structure, it only focuses on the relations between the coordinates, while other low-level geometric features such as normal and curvature are not considered. 
(2) Second, compared to the point coordinates, the proposed rotation-invariant features will inevitably lose geometric information.
In the future, we might explore the possibility of using other modalities such as images to break the current accuracy bottleneck of 3D object detection.

\bibliography{aaai25}

\end{document}